\documentclass{article}
\usepackage{spconf,amsmath,graphicx,color}
\usepackage{subcaption,multirow,arydshln}

\title{On the use of Self-supervised Pre-trained Acoustic and Linguistic Features for Continuous Speech Emotion Recognition}

\name
 {Manon Macary$^{1,2}$, Marie Tahon$^{1}$, Yannick Estève$^{3}$, Anthony Rousseau$^{2}$}
 \address
	{$^{1}$ LIUM, Le Mans Université, France\\
	$^{2}$ Allo-Media\\
	$^{3}$ LIA, Avignon Université, France
	}

\begin{document}
%
\maketitle
\begin{abstract}

Pre-training for feature extraction is an increasingly studied approach to get better continuous representations of audio and text content.
In the present work, we use wav2vec and camemBERT as self-supervised learned models to represent our data in order to perform continuous emotion recognition from speech (SER) on AlloSat, a large French emotional database describing the satisfaction dimension, and on the state of the art corpus SEWA focusing on valence, arousal and liking dimensions. 
To the authors’ knowledge, this paper presents the first study showing that the joint use of wav2vec and BERT-like pre-trained features is very relevant to deal with continuous SER task, usually characterized by a small amount of labeled training data. Evaluated by the well-known concordance correlation coefficient (CCC), our experiments show that we can reach a CCC value of 0.825 instead of 0.592 when using MFCC in conjunction with word2vec word embedding on the AlloSat dataset.
\end{abstract}
\section{Introduction}
\label{sec:intro}
    Human-to-human dialogue is a great source of interest for academic researchers and companies. 
    It is important to understand and to be understood by other people to maintain a functional society.
    One of the domains where understanding is necessary is in commercial conversations and particularly in customer services.
    With the democratization of phones, call centers are a growing service offered by companies, and it is today one of the most used mean of direct communication with their customers.
    Within this context, recognize both customer satisfaction and frustration emotional states in call center conversations is crucial to customer services, in order to help companies to improve these services.
    Human machine interactions can also benefit from satisfaction and frustration recognition, in order to make the dialog more natural between humans and machines.

    Affective computing is a field of research at the crossroads between artificial intelligence and human emotions. Since decades, different theories have been used to encode human emotion into machine readable labels. 
    Among these theories, discrete emotional labels such as the ``Big Six''~\cite{Ekman1999} (i.e. anger, joy, sadness, disgust, fear and surprise) often added with a ``neutral'' class have been extensively used in emotion classification as they consider only a limited number of emotions.
    One drawback of this theory is that very few emotions are discretized, so all the range of human behaviour can not be captured.
    Another theory, which suppress part of this limitation, allows to capture more subtle nuances by characterizing emotions with continuous dimensions, notably activation (passive/active) and valence (positive/negative)~\cite{Russel1997}, but also dominance (weak/strong self-control), expectation~\cite{Scherer2005}, or conductive/obstructive axis.
    Other dimensions have been studied to better fit with the application context. For example, the satisfaction is used for call-center conversations study~\cite{LRECdeNous}, and liking have been predicted jointly with activation and valence in dyadic conversations~\cite{SEWA}.
    
    It is usually established~\cite{Devillers2006ReallifeED} that emotion in speech can be transmitted through linguistic messages conveyed by words and paralinguistic ones conveyed by the acoustic signal.
    While activation is known to be well-recognized from acoustic features, valence is known to be better recognized with linguistic ones~\cite{FEATURES_EMO_DETECT}.
    According to Scherer~\cite{Scherer2005}, satisfaction and frustration can be considered as combination of activation and valence dimensions. Consequently, linguistic and acoustic features should both be highly relevant to retrieve such dimensions. For these reasons, we decided to study the impact of linguistic and acoustic features extracted with self-supervised pre-trained models in order to predict continuous emotion.

\section{Related works}
\label{sec:motivations}

    Because manual discrete and continuous emotion annotation is a highly subjective perception task, it has to be done by multiple annotators to be relevant, and this explains the huge cost of the creation of large emotional speech datasets. Consequently, Speech Emotion Recognition (SER) databases are usually quite small (SEWA~\cite{SEWA}: $\sim$44~h, RECOLA~\cite{Ringeval2013IntroducingTR}: $\sim$2.8~h, AlloSat~\cite{LRECdeNous}: $\sim$37~h).
    That is one of the reasons why deep neural networks (DNN) have been used only recently in SER in comparison to automatic speech recognition (ASR) where accessible databases are drastically bigger (\textsl{e.g.} TED-LIUM~3~\cite{Hernandez_2018}: $\sim$450~h, LibriSpeech~\cite{librispeech}:  $\sim$960~h).
    
    Transfer learning~\cite{goodfellow2016deep} within the deep learning paradigm is a machine learning method where a neural network trained for a task is reused partially or entirely as the starting point to train or fine-tune a neural network on a second task. 
    Such methods were also proposed to limit the impact of lack of data when only small databases are available to train a neural network for a specific domain or task. 
    Transfer learning is widely used in Sentiment Analysis, where large databases are used to train generic cues which are fed into the training process, leading to better generalization abilities given limited training data~\cite{dong-de-melo-2018-helping}.
    
    As a variant of transfer learning, self-supervised learning of speech or language representations has been proposed these last few years, for instance with the BERT system~\cite{devlin-etal-2019-bert}, used for textual representation. 
    Such representations, computed by neural models trained on huge amounts of unlabeled data, have  shown their effectiveness on some tasks under certain conditions, for instance for computer vision~\cite{Nanni2017HandcraftedVN} and Natural Language Processing (NLP) tasks such as 
    Single-Sentence Classification, Text Similarity, Relevance Ranking~\cite{Liu2019MultiTaskDN,young2018review,yang2019xlnet}, ASR~\cite{Kahn2020Libri,Liu2020Mockingjay}, or speech translation
   ~\cite{nguyen2020investigating}.
    
    In this paper, we investigate, for SER tasks, the use of such speech and/or textual representations computed by models pre-trained through self-supervised learning.
    Since these already existing pre-trained models were initially designed for speech recognition ASR~\cite{Schneider2019wav2vecUP} or natural language understanding~\cite{devlin-etal-2019-bert}, it is not obvious that they are also relevant for SER. For instance, at the acoustic level ASR tends to focus on phone level that lasts about 30~ms while emotions are usually supported on about 1~s of speech.
    
     In the remainder, we introduce the pre-trained models we used to extract acoustic and linguistic features in  Section~\ref{sec:pretrain}. The datasets and the extracted features are presented in Section~\ref{sec:data}. SER experiments are explained in Section~\ref{sec:exp} and results are presented in Section~\ref{sec:result} followed by the conclusion (Section ~\ref{sec:conclusion}).

\section{Self supervised linguistic and speech continuous representation}
\label{sec:pretrain}

    As seen above, the use of pre-trained models trained through self-supervised learning to extract continuous features has shown its utility in a lot of domains.
    In this work, we focus on linguistic and acoustic continuous representations extracted thanks to specific pre-trained models.
    
    \subsection{Linguistic representation}
    Word embedding is one of the most popular continuous representation of textual data. These embeddings are vector representations of a particular word. 
    Among them, word2vec~\cite{word2vec} is one of the most used in Sentiment Analysis tasks such as Polarity or Emotion State Classification~\cite{Rodrigo2020,dong-de-melo-2018-helping}.
    Word2vec word embeddings are static: a word has always the same vector representation, 
    regardless of the true meaning of the word and the context of its occurrence. 
    This is problematic for polysemic words, for instance frequent in French~\cite{pustejovsky1996lexical}.
    
    Other models have been released these last two years, such as BERT~\cite{devlin-etal-2019-bert} (trained on English data then multi-lingual data) or Ernie~\cite{zhang2019ernie}, which are language and context dependent. These models need a lot of data to be trained, such as the entire Wikipedia totalling over 2,500 million words and Book Corpus totalling over 800 million words. 
    
    In this study, linguistic representation will be used to process French data.
    Very recently, CamemBERT~\cite{martin-etal-2020-camembert}, FlauBERT~\cite{le2020flaubert} and GermanBERT~\footnote{https://deepset.ai/german-bert} models were released for French and German while Ernie models are only available in Chinese and English. 
    %
    As far as the authors know, this is the first time that such models are used for feature extraction in order to perform SER tasks.
    
    \subsection{Acoustic representation}
    In SER, finding the better acoustic feature set is still an ongoing active sub-field of research in the domain~\cite{Jing2018ProminenceFE}. 
    Most of the handcrafted sets~\cite{EGMAPS,schuller2013interspeech} intend to describe prosody in the signal, with low level descriptors (LLDs) capturing intensity, intonation, rhythm or voice quality. 
    Another approach is to extract spectral features: mel frequency cepstral coefficients (MFCCs) are clearly the most often used as they are robust to noisy signals even if they have not been designed for prosody nor emotion.
    Recently wav2vec~\cite{Schneider2019wav2vecUP} and Audio AlBERT~\cite{chi2020audio} were introduced in ASR and speaker identification as one of the first pre-trained approaches to extract context dependent features from raw signals for ASR tasks.
    In our study, we propose to investigate the use of a pre-trained wav2vec model to extract acoustic features in order to retrieve the continuous emotional state of the speaker.

\section{Datasets and Feature Extraction}
\label{sec:data}
In order to analyze the satisfaction and frustration, we chose to use the AlloSat and SEWA corpus described below.
    
    \subsection{Speech Datasets}
        Both AlloSat and SEWA datasets are annotated to analyze continuous emotion in speech. Table~\ref{sumUpAlloSat} summarizes their main characteristics.
        AlloSat~\cite{LRECdeNous} is a French real-life call-center corpus composed of 303 conversations as indicated in Table~\ref{sumUpAlloSat}. More precisely, adults callers (i.e. customer) asking for various information such as contract information, global details on the company, complains directed towards various domains organizations (energy, travel agency, real estate agency, insurance, ...). All conversations were annotated by three annotators in a continuous manner along the satisfaction dimension (a dimension going from frustration to satisfaction with a neutral state in the middle) sampled every 250~ms. 
        It permits the analyze of both satisfaction and frustration in real-life in knowingly stressful environment. 
        As it consists on phone conversations, the audio is originally sampled at 8kHz.
        This corpus has the advantage to be speaker-independent, domain-independent and composed of real-life recording.
        It is divided into three subsets: the Train set containing 201 conversations corresponding to $\sim$25h of audio signal and $\sim$16h of speech; the Development set composed of 42 conversations and the Test set containing 60 conversations. 
        Both Development and Test sets are composed of ~6h of audio signal and ~3h of speech.
        
        To extend our work to another accessible dataset, we chose to also use a part of the SEWA dataset~\cite{SEWA}.
        We used the German set of the cross-cultural Emotion Database SEWA that is composed of 24 audiovisual recordings of dyadic conversations as showed by Table~\ref{sumUpAlloSat}, sampled at 16kHz.
        Pairs are discussing for less than 3 minutes about an advert seen beforehand.
        As they are two speakers per document, Kossaifi et al.~\cite{SEWA} chose to duplicate each document and to add a flag denoting the current speaker, therefore there are 48 documents in the database.
        The SEWA corpus is now a reference in the community, as it has been used in the two last Audio/Visual Emotion Challenges (AVEC)~\cite{AVEC,AVEC2019}. 
        A continuous annotation among three dimensions (arousal, valence and liking) was made by six annotators and sampled every 100~ms. 
        The liking axis describes how much the subjects liked the advert.
        The corpus is divided into two subsets: the Train set containing 34 documents (i.e. 17 conversations) and and the Development set containing 14 documents (i.e. 7 conversations). Test gold references are not distributed.
        Annotations were merged to form a gold reference for both datasets, used in the prediction task.
        
        \begin{table}[ht!]
        \centering
        \begin{tabular}{lll}
            \hline \textbf{Statistics} & \textbf{AlloSat} &\textbf{SEWA} \\ 
            \hline
             language                   &French &German \\
             number of conversations	&303	&24 \\
             number of speakers      &308   &48 \\
            \hline
            total duration      	&37h23'27'' &1h05'10''\\
            \hline
            min duration conversations		&32'' &47''		\\
            max duration conversations		&41'11'' &2'55''		\\
            mean duration conversations		&7'24'' &2'44''    \\
            \hline
            number of annotators        &3      &6 \\
            emotional segment duration  & 250~ms    & 100~ms\\
            \hline
        \end{tabular}
        \caption{\label{sumUpAlloSat} Summary of AlloSat and SEWA characteristics}
        \end{table}

    \subsection{Acoustic features}
    \label{ssec:input_acoustic}
        We build both baseline and `pre-trained' features\footnote{For the sake of simplicity, in the remainder pre-trained features will refer to features computed by models pre-trained through self-supervision} for the acoustic modality in order to compare them.
        In the following, an emotional segment lasts 250~ms in AlloSat and 100~ms in SEWA, and corresponds to the annotation timestep.
        \vspace{-0.2cm}
        \subsubsection{Baseline features}
            We use both eGeMAPS and MFCC features to build the baseline features used in our previous work~\cite{macary2020}.
            In the MFCC set, 24 MFCCs are extracted each 10~ms on a 30~ms window and summarized with mean and standard deviation over the emotional segment (250
           ~ms for AlloSat and 100~ms for SEWA) using librosa toolkit~\footnote{https://librosa.github.io/librosa/}. 
           In the eGeMAPS set, we extract 23 LLDs~\cite{EGMAPS}. Mean and standard deviation of these 23 LLDs are computed over the emotional segment. This feature set is extracted with the toolkit OpenSmile~\cite{OPEN_SMILE}. An additional binary flag denoting speaker presence extracted from speech turns, 
           is added to acoustic feature sets extracted on SEWA only.
        \vspace{-0.2cm}
        \subsubsection{Pre-trained features}
            Wav2vec~\cite{Schneider2019wav2vecUP} is a pre-trained model through self-supervision: it is learned to predict future samples from a current window analysis. This model is composed of two distinct convolutional neural networks. A first `encoder network' converts the audio signal into a new representation that is given to the second network, the `context network', which takes care of the context by aggregating multiple time step representation in order to use a receptive field of about 210~ms. Both are then used to compute a contrastive loss function. The resulting embedding is a 512-dimensional feature vector.
            
            We use two pre-trained models. The first one, called `wav2vec-EN' is provided by Schneider et al. in ~\cite{Schneider2019wav2vecUP}, trained on Librispeech corpus~\cite{librispeech} consisting of 960 hours of English audio book. 
            We also train our own model, called `wav2vec-FR' on French unlabeled call center conversations on over 500 hours of private-owned data.
            Unfortunately, since no large conversational German database was at our disposal, we could not build a German model.
            Features were extracted on an upsampled\footnote{We used FFMpeg resampling function with \textit{sinc} interpolation function} 
            Allosat (from 8 to 16kHz) using both `wav2vec-FR' and `wav2vec-EN' and on the SEWA dataset using `wav2vec-EN'. 
            In the end, each emotional segment is represented by a 512-dimensional vector which consists of the averaged values of obtained embeddings over the emotional segment.
    
    \subsection{Linguistic features}
    \label{ssec:input_linguistic}
        As we do not have the transcription of the SEWA corpus, the linguistic modality is only explored with the AlloSat corpus, for which only automatic transcriptions are available.
    
        From the conversation transcriptions, we build the linguistic features. Continuous prediction needs alignment between the linguistic features and the annotation timescale. 
        As we face a sequence problem, as we have to respect the timescale, we align the transcription to the annotation. We faced many issues: one word on multiple emotional segments (longer than the annotation window or beginning at the end of one and continue on the following), multiple words in the same emotional segment and stop words (such as a, the, of,...). Therefore we define a protocol to align transcription and annotation as following:
        \vspace{-0.2cm}
            \begin{itemize}
                \item All words are kept, even the stop words.
                \vspace{-0.2cm}
                \item If one word is pronounced over multiple consecutive emotional segments, its embedding vector is duplicated on each corresponding emotional segment.
                \vspace{-0.2cm}
                \item If multiple words are pronounced in one emotional segment, embeddings of each words are averaged.
            \end{itemize}
        
        \subsubsection{Baseline features}
            Baseline linguistic aspect is modeled with word2vec. The system has been trained with the toolkit GENSIM~\cite{gensim}, using private data owned by Allo-Media composed of manual call transcriptions received by call centers, totaling over 4.5 million words. 
            The embedding size has been set to 40 due to preliminary works dealing with feature fusion.
    
        \subsubsection{Pre-trained features}
            Inspired by RoBERTa~\cite{liu2019roberta} and BERT, CamemBERT is a multi-layer bidirectional Transformer as described by Martin et al. in~\cite{martin-etal-2020-camembert}.
            CamemBERT is trained on the Masked Language Modeling (MLM) task which consists of replacing some tokens by either the token $<$MASK$>$ or a random token and asking the model to correct the tokens. The network uses a cross-entropy loss. The input consists of a mix of whole words and sub-words in order to take advantage of the context.
            
            We use the pre-trained model, called `camemBERT-base', trained on the French part of the OSCAR corpus~\cite{ortizsuarez2019} consisting of a set of monolingual corpora extracted from Common Crawl snapshot and totaling 138GB of raw text and 32.7B tokens after sub-word tokenization.
            Features were extracted on Allosat by using this pre-trained model, and we summarized the results by averaging the continuous representations of sub-words occurring in the current emotional segment. 
            In total, we use a 768-dimensional feature vector.

        \begin{figure}
            \centering
             \includegraphics[width=.8\linewidth]{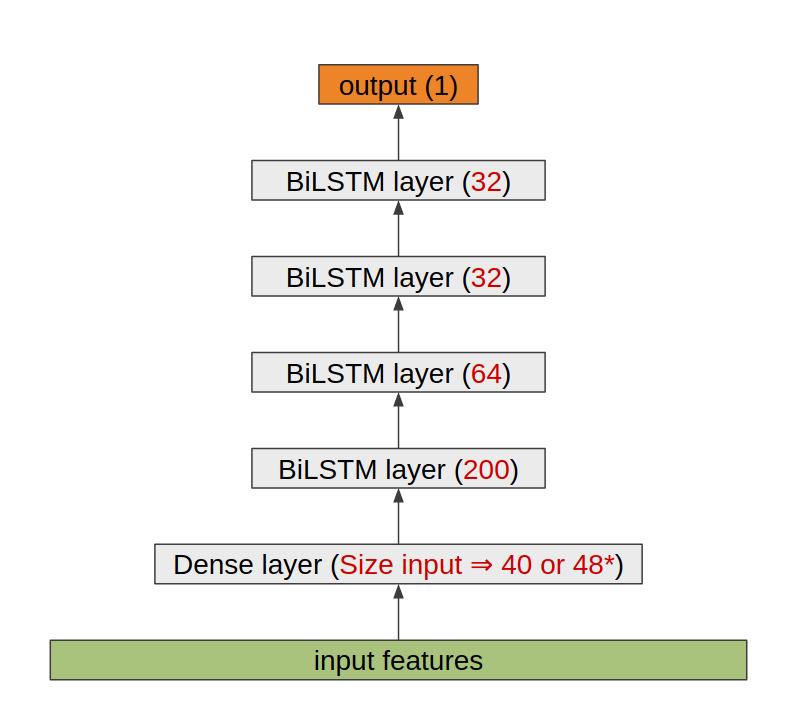} 
            \caption{Description of the model: units are written in red. The first dense layer is optional and not used in with baseline features as it is used to reduce the input size of pre-trained features. * `Size input' for acoustic and linguistic features is respectively set to 512 and 768.}
            \label{fig:models}
        \end{figure}

                \begin{table}[ht]
        \centering
            \begin{tabular}{ll|lll}
                \hline 
                 &  &\textbf{Arousal} &\textbf{Valence} &\textbf{Liking} \\
                \textbf{Modality} &\textbf{\# input} 
                &\textbf{Dev} &\textbf{Dev} &\textbf{Dev} \\
                \hline
                AUDIO \\
                \hline
                MFCC            &49 &0.501 &0.452 &0.133 \\
                eGeMAPS         &47 &0.479 &0.375 &0.197 \\
                wav2vec-EN      &513 &0.251 &0.215 &0.254 \\
                \hline
                \end{tabular}
                \caption{\label{tab:resultSEWA} Results of the audio modality in terms of CCC computed on Development sets on the SEWA dataset. Experiments were made with an additional binary flag denoting speaker presence extracted from speech turns.}
        \end{table}

        \begin{table}[ht]
        \centering
            \begin{tabular}{ll|ll}
                \hline 
                & &\multicolumn{2}{c}{\textbf{Satisfaction}} \\
                \textbf{Modality} &\textbf{\# input} &\textbf{Dev} &\textbf{Test} \\
                \hline
                AUDIO \\
                \hline
                MFCC            &48 &0.698 &0.513 \\
                eGeMAPS         &46 &0.422 &0.354 \\
                wav2vec-EN      &512 &0.851 &\textbf{0.730} \\
                wav2vec-FR      &512 &\textbf{0.865} &0.635 \\
                \hline
                TEXT \\
                \hline
                Word2vec        &40 &0.805 &0.569 \\ 
                CamemBERT       &768 &\textbf{0.896} &\textbf{0.799} \\
                \hline
                DECISION FUSION \\
                \hline
                MFCC + word2vec  &48 + 40 &0.838 &0.659\\
                wav2vec-EN + word2vec  &512 + 40 &0.878 &0.750\\
                wav2vec-EN + camemBERT &512 + 768 &\textbf{0.907}   &\textbf{0.825} \\
                \hline
            \end{tabular}
        \caption{\label{tab:resultFinal} Comparison of the audio and text modalities, and decision fusion results in terms of CCC computed on Development and Test sets on AlloSat. Fusion results are given together with optimized weights (on Dev): 0.38 MFCC + 0.62 word2vec, 0.63 wav2vec-EN + 0.37 word2vec and 0.31 wav2vec-EN + 0.69 camemBERT.}
        \end{table}

\section{Speech Emotion Recognition experimental setup}
\label{sec:exp}
    In order to highlight how relevant pre-trained features are for continuous SER,
    we propose to compare the prediction results with pre-trained features (wav2vec for audio and camemBERT for text) and baseline features (MFCC for audio and word2vec for text) on AlloSat (text and audio) and SEWA (audio only).
    
    \subsection{Fusion of the modalities}
        
        We decided to run experiments using audio and text modalities independently. Each experiment corresponds to a different set of features in input. Four acoustic feature sets and two linguistic feature sets, presented in Section~\ref{ssec:input_acoustic} and ~\ref{ssec:input_linguistic}, are tested. As presented in the introduction, both modalities are relevant for emotion recognition, that is why we also present fusion results. 
        
        Different fusion protocols were explored: feature level fusion (feature vector concatenation), model level fusion (networks layer concatenation) and decision level fusion. Previous results on the concatenation of our baseline feature vectors (MFCC + word2vec) and of the first layer of acoustic and linguistic models have shown poor results. Therefore only decision fusion results have been experimented with pre-trained features.
        More precisely, the decision fusion final prediction consists in the weighted average of each modality prediction.
        We experiment on the weight to give to each modality from 0.1 to 0.9 with a step of 0.01 for each fusion experiments.
        
    \subsection{SER module}
        
        A deep Neural Network (DNN) is used for the prediction task using bidirectionnal Long Short-Term Memory (biLSTM). This architecture inspired from~\cite{basepaper} is composed of 4 biLSTM layers composed of respectively 200, 64, 32, 32 units with a tanh activation. Experiments have been carried out on the number of layers (from 3 to 6) and on the number of units per layer and this architecture achieved better results. A single output neuron is also used to predict the regression every emotional segments (250~ms or 100~ms). Therefore we learned 4 models for the acoustic modality (one for each dimension: satisfaction, arousal, valence and liking). Neither dropout nor batch normalisation is used in this approach.
        
        In order to compare baseline and pre-trained features, we add an optional dense layer to the network, after the input layer, in case of input pre-trained features. 
        This aims at reducing the input size to baseline feature dimensions: 40 for audio and 48 for text, as shown in the Figure~\ref{fig:models}. Additional experiments on the baseline systems have also been done with a dense layer in input to ensure that the results were not improved only thanks to a higher number of parameters: they did not show any significant improvement for our task. A mean and variance normalization of the input features is done for all experiments.
    
         \begin{figure*}
            \centering
            \begin{subfigure}{.49\textwidth}
              \centering
              \includegraphics[width=.99\linewidth]{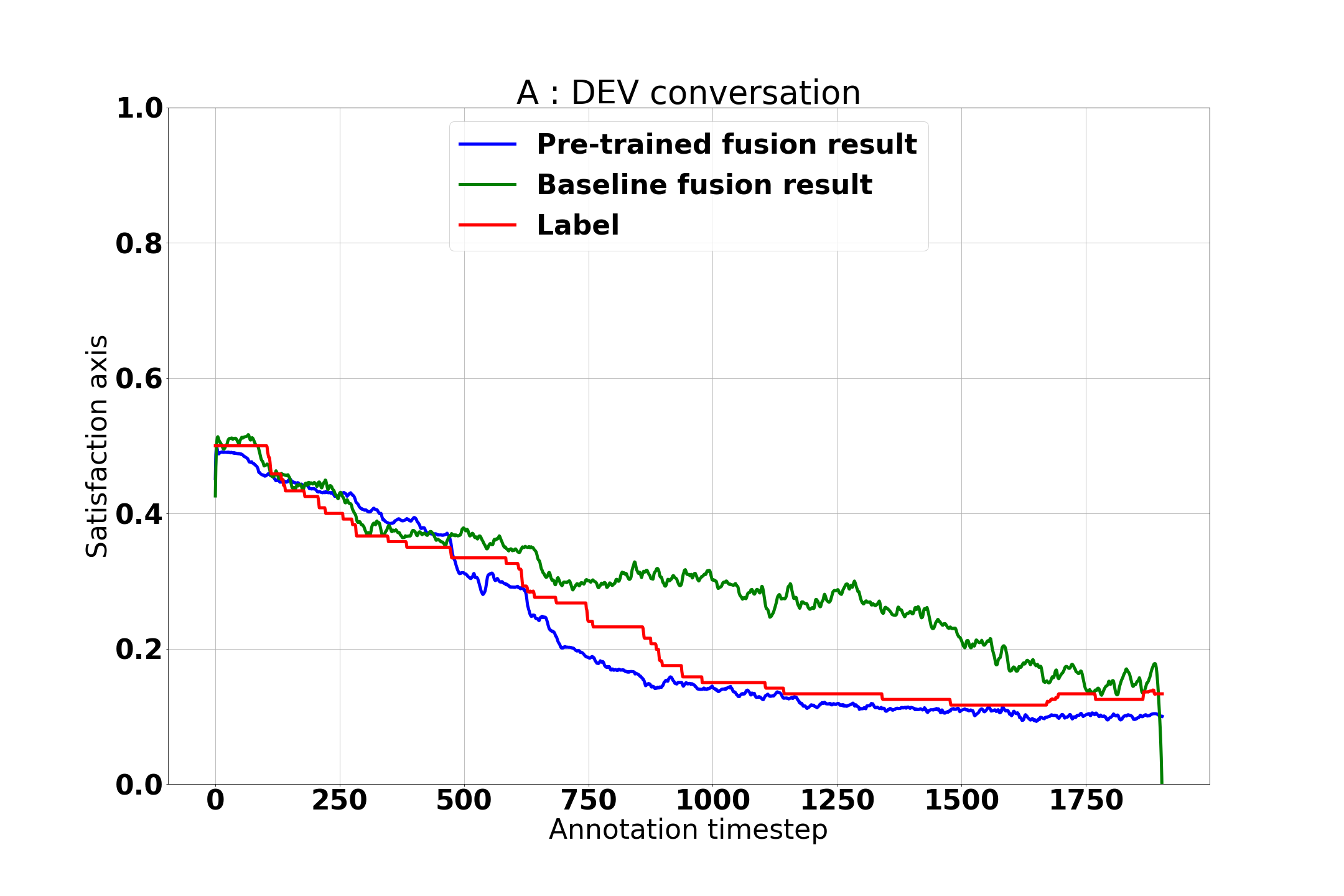}  
            \end{subfigure}
            \begin{subfigure}{.49\textwidth}
              \centering
              \includegraphics[width=.99\linewidth]{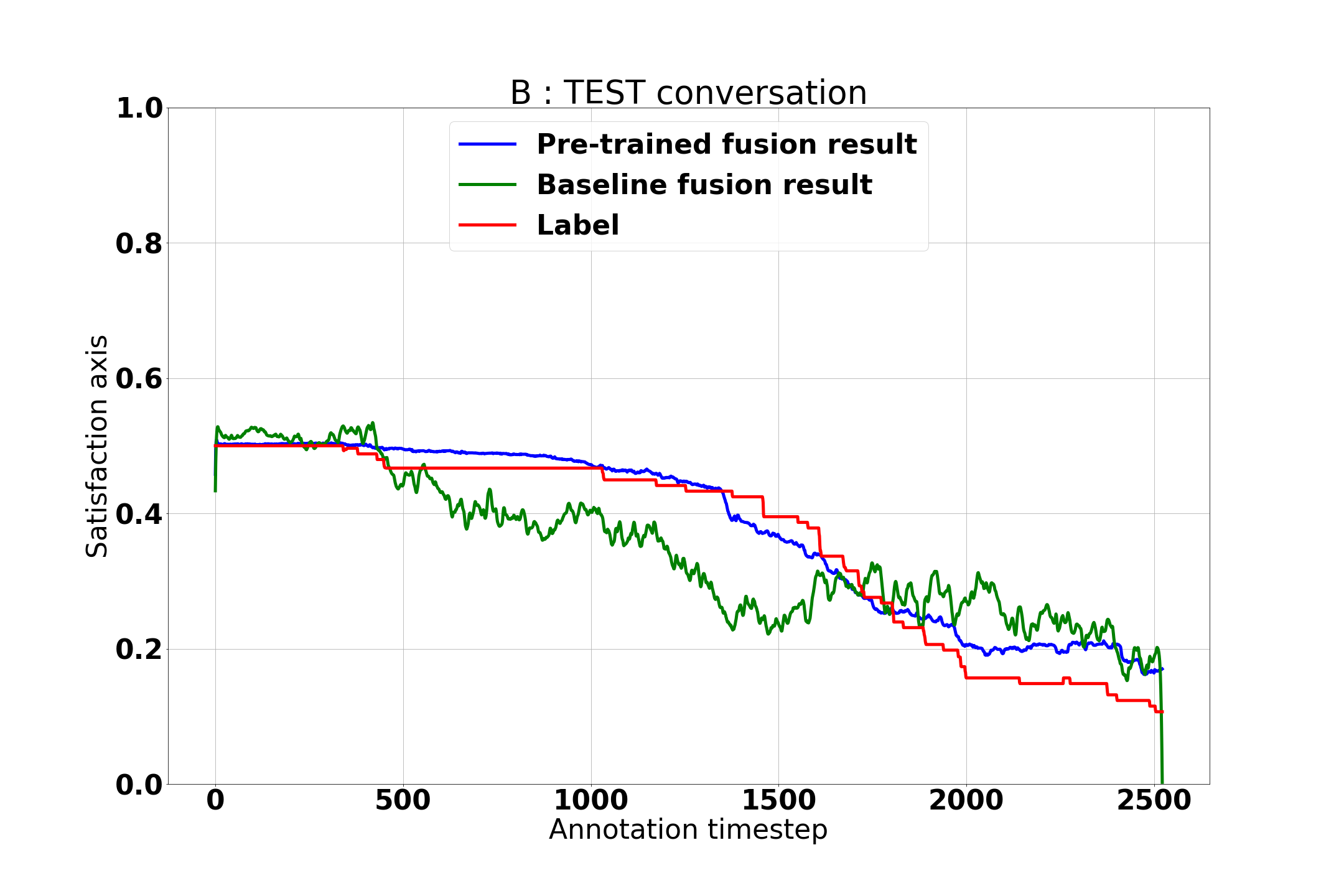}  
            \end{subfigure}
            \caption{Evolution of reference satisfaction (red) and the two modalities fusion : baseline MFCC+word2vec (green) called $X_b$ and pre-trained wav2vec+camemBERT (blue) called $X_p$ in two conversations from AlloSat dev (A) and test (B) subsets. $ccc(A_p) = 0.971$; $ccc(A_b) = 0.748$, $ccc(B_p) = 0.972$ ; $ccc(B_b) = 0.767$.}
            \label{fig:two_predictions}
        \end{figure*}

    \subsection{Loss and evaluation function}
        The concordance correlation coefficient (CCC)~\cite{CCC} is computed as the loss function for training the networks, as it is widely used in the retrieval of continuous emotions.
        CCC score goes from 0 (chance level) to 1 (perfect) and is calculated according to the eq.~\ref{eq:CCC_score}, where $x$ correspond to the predicted value and $y$ the label. $\mu_x$ and $\mu_y$ are the means for the two variables and $\sigma_x$ and $\sigma_y$  are the corresponding variances. $\rho$ is the correlation coefficient between the two variables $\sigma_x$ and $\sigma_y$ therefore the covariance coefficient.
        
        \begin{equation}
            CCC = \frac{2\rho\sigma_x\sigma_y}{\sigma_x^2 + \sigma_y^2 + (\mu_x - \mu_y)^2}
        \label{eq:CCC_score}
        \end{equation}
        CCC is also used as the evaluation metric. 
    
    \subsection{Hyper-parameters}
        The DNNs are implemented with the pytorch framework~\cite{pytorch}. Preliminary experiments were done on the development set, in order to settle the network architecture and the following hyper-parameters.
        Training is done on batches of 15 conversations using the Adam optimiser.
        The learning rate is optimized at 0.001 by empirical method, tested on a range from 0.001 to 0.02 by a 0.005 step. 
        After preliminary experiments, networks were not improving after the first 400 epochs, so the total number of epochs is set to 500. 
        The Test set is evaluated using the networks which have the best scores on the Development set.

\section{Discussion}
\label{sec:result}
    
    All results are reported in Table~\ref{tab:resultSEWA} and Table~\ref{tab:resultFinal}.
    We can observe that pre-trained features are achieving awesome results in comparison to baseline features, both in acoustic and linguistic aspects on the AlloSat corpus.
    The improvement obtained on acoustics is very important on AlloSat: from 0.513 of CCC obtained with MFCCs features, we reach 0.730 with wav2vec-EN.
    The improvement obtained on linguistic modality is less spectacular than the one obtained on acoustics, but seems significant: we reach 0.750 of CCC value with word2vec while we get 0.825 with CamemBERT when fused with wav2vec.
    It is not easy, without more investigations, to explain why we get better results with CamemBERT representations trained on generic data while we trained the word2vec representations on in-domain data. 
    A lot of differences exist by nature between CamemBERT and word2vec (complexity of the neural architecture, context-dependent dynamic embeddings \textsl{vs.} static embeddings, sub-words vs words...). 
    The computation of the CamemBERT needs a lot of GPUs, data and time. Fortunately this model is available and this makes possible to get very good results on the targeted SER tasks without strong efforts and resources.
    Surprisingly, `wav2vec-FR' performs better than `wav2vec-EN' on the Dev set but not on the Test set. This can be explained by the smaller amount (460~h) of data used to train the French wav2vec model, which is half of the one used in `wav2vec-EN' (almost 1000~h). 
    Notice that we also found out that the test set is a harder set than the dev one. To prove this point, we did experiments where we switched the two sets. 
    
    However the results using pre-trained features on SEWA are very disappointing and classical handcrafted features still do a better job on activation and valence dimensions. Anyway, we can notice that pre-trained features are interesting to retrieve the liking dimension. Our hypothesis remains in the data amount difference between the two corpus: AlloSat Train set is composed of 25 hours of audio recording while SEWA Train set have less than 2 hours of conversations. Despite the use of pre-trained features, the network is not able to achieve good results, probably because of the small amount of data.
    
    Additional results, which are not presented here, show that the addition of a binary feature denoting the presence of the speaker does not improve the results on AlloSat (the agent speech has been removed and replaced by a jazzy sound) but does improve on SEWA (both speakers voices are present).
    The compared performances obtained with audio modality shows that deeper investigations are required to understand why pre-trained features fits best with the French call-center corpus annotated in satisfaction and frustration and not with German dyadic conversations. 
    
    Results on the fusion indicates that acoustic and linguistic information are both useful in order to continuously predict the satisfaction dimension. This confirms the multimodal aspect of real-life emotions.
    On Figure~\ref{fig:two_predictions}, we can see that the pre-trained fusion (wav2vec + camemBERT) result (blue) is closer to original labels (red) than baseline fusion (MFCC + word2vec) result (green). We can also notice that pre-trained features predict a smoother curve than baseline features, probably because MFCCs are not able to take a large context into account while pre-trained features do. To the authors' knowledge, it is the first time continuous emotion predictions are presented graphically and not only in terms of CCC.

\section{Conclusion}
\label{sec:conclusion}

    In this article, we showed that continuous representations computed by available pre-trained models can be successfully used for continuous speech emotion recognition using two modalities: acoustic (features extracted from the raw signal) and linguistic (features extracted from the automatic transcription of the conversations).
    Compared to a set of baseline features which achieved state-of-art results on the AlloSat corpus to the best of our knowledge, we improve the results on both modalities to respectively +0.217 and +0.230 CCC scores and outperforms the previous state-of-art.
    With the use of pre-trained models, we benefit from features learnt on very large databases, while emotional databases are generally small. Moreover, our experiments indicate that the language does not seem to be restrictive.
    However, this kind of approach does not generalize to activation and valence from SEWA corpus. We hypothesize that it is mainly due to the small amount of data in the training set.
    %
    We also showed that acoustic and linguistic predictions can be fused altogether to better model the emotional state of a speaker regarding satisfaction and frustration, improving the best mono-modality result from 0.799 to 0.825 CCC scores.


\bibliographystyle{IEEEbib}
\bibliography{bib_light}

\begin{thebibliography}{10}

\bibitem{Ekman1999}
P.~Ekman,
\newblock {\em Basic Emotions}, pp. 301--320,
\newblock Wiley, New-York, 1999.

\bibitem{Russel1997}
J.~Russel,
\newblock {\em Reading emotions from and into faces: Resurrecting a
  dimensional-contextual perspective}, pp. 295--360,
\newblock Cambridge University Press, U.K., 1997.

\bibitem{Scherer2005}
K.~R. Scherer,
\newblock ``What are emotions? and how can they be measured?,''
\newblock {\em Social science information}, vol. 44, no. 4, pp. 695--729, 2005.

\bibitem{LRECdeNous}
M.~Macary, M.~Tahon, Y.~Est{\`e}ve, and A.~Rousseau,
\newblock ``{AlloSat}: A new call center french corpus for satisfaction and
  frustration analysis,''
\newblock in {\em Proc. of Language Resources and Evaluation Conference
  (LREC)}, Virtual Conference, 2020, pp. 1590--1597.

\bibitem{SEWA}
J.~Kossaifi, R.~Walecki, Y.~Panagakis, J.~Shen, M.~Schmitt, et~al.,
\newblock ``{SEWA DB}: A rich database for audio-visual emotion and sentiment
  research in the wild,''
\newblock {\em IEEE transactions on pattern analysis and machine intelligence},
  pp. 1--1, 2019.

\bibitem{Devillers2006ReallifeED}
L.~Devillers and L.~Vidrascu,
\newblock ``Real-life emotions detection with lexical and paralinguistic cues
  on human-human call center dialogs,''
\newblock in {\em Proc. of INTERSPEECH}, Pittsburgh, Pennsylvanie, USA, 2006,
  pp. 801--804.

\bibitem{FEATURES_EMO_DETECT}
Y.~Alva~M, N.~Muthuraman, and J.~Paulose,
\newblock ``A comprehensive survey on features and methods for speech emotion
  detection,''
\newblock in {\em Proc. of IEEE International Conference on Electrical,
  Computer and Communication Technologies (ICECCT)}, Coimbatore, Tamilnadu,
  India, 2015, pp. 1--6.

\bibitem{Ringeval2013IntroducingTR}
F.~Ringeval, An. Sonderegger, J.~S. Sauer, and D.~Lalanne,
\newblock ``Introducing the {RECOLA} multimodal corpus of remote collaborative
  and affective interactions,''
\newblock {\em 10th IEEE International Conference and Workshops on Automatic
  Face and Gesture Recognition (FG)}, pp. 1--8, 2013.

\bibitem{Hernandez_2018}
F.~Hernandez, V.~Nguyen, S.~Ghannay, N.~Tomashenko, and Y.~Estève,
\newblock ``{TED-LIUM} 3: Twice as much data and corpus repartition for
  experiments on speaker adaptation,''
\newblock in {\em Proc. of Conference on Speech and Computer (SPECOM)},
  Leipzig, Germany, 2018, pp. 198--208.

\bibitem{librispeech}
V.~Panayotov, G.~Chen, D.~Povey, and S.~Khudanpur,
\newblock ``Librispeech: An {ASR} corpus based on public domain audio books,''
\newblock in {\em Proc. of ICASSP}, South Brisbane, Queensland, Australia,
  2015, pp. 5206--5210.

\bibitem{goodfellow2016deep}
I.~Goodfellow, Y.~Bengio, and A.~Courville,
\newblock {\em Deep learning},
\newblock MIT press, 2016.

\bibitem{dong-de-melo-2018-helping}
X.~Dong and G.~de~Melo,
\newblock ``A helping hand: Transfer learning for deep sentiment analysis,''
\newblock in {\em Proc. of ACL}, Melbourne, Australia, 2018, pp. 2524--2534.

\bibitem{devlin-etal-2019-bert}
J.~Devlin, M.~Chang, K.~Lee, and K.~Toutanova,
\newblock ``{BERT}: Pre-training of deep bidirectional transformers for
  language understanding,''
\newblock in {\em Proc. of the North American Chapter of the Association for
  Computational Linguistics (NAACL)}, Minneapolis, Minnesota, USA, 2019, pp.
  4171--4186.

\bibitem{Nanni2017HandcraftedVN}
L.~Nanni, S.~Ghidoni, and S.~Brahnam,
\newblock ``Handcrafted vs. non-handcrafted features for computer vision
  classification,''
\newblock {\em Pattern Recognition}, vol. 71, pp. 158--172, 2017.

\bibitem{Liu2019MultiTaskDN}
X.~Liu, P.~He, W.~Chen, and J.~Gao,
\newblock ``Multi-task deep neural networks for natural language
  understanding,''
\newblock in {\em Proc. of ACL}, Florence, Italy, 2019, pp. 4487--4496.

\bibitem{young2018review}
T.~Young, D.~Hazarika, S.~Poria, and E.~Cambria,
\newblock ``Recent trends in deep learning based natural language processing
  [review article],''
\newblock {\em IEEE Computational Intelligence Magazine}, vol. 13, no. 3, pp.
  55--75, 2018.

\bibitem{yang2019xlnet}
Z.~Yang, Z.~Dai, Y.~Yang, J.~Carbonell, R.~R Salakhutdinov, et~al.,
\newblock ``{XLNet}: Generalized autoregressive pretraining for language
  understanding,''
\newblock in {\em Advances in Neural Information Processing Systems (NIPS)},
  Vancouver, Canada, 2019, pp. 5753--5763.

\bibitem{Kahn2020Libri}
J.~Kahn, M.~Rivière, W.~Zheng, E.~Kharitonov, Q.~Xu, et~al.,
\newblock ``Libri-light: A benchmark for {ASR} with limited or no
  supervision,''
\newblock in {\em Proc. of ICASSP}, Virtual Conference, 2020, pp. 7669--7673.

\bibitem{Liu2020Mockingjay}
A.~T. Liu, S.~Yang, P.~Chi, P.~Hsu, and H.~Lee,
\newblock ``Mockingjay: Unsupervised speech representation learning with deep
  bidirectional transformer encoders,''
\newblock in {\em Proc. of ICASSP}, Virtual Conference, 2020, pp. 6419--6423.

\bibitem{nguyen2020investigating}
H.~Nguyen, F.~Bougares, N.~Tomashenko, Y.~Est{\`e}ve, et~al.,
\newblock ``Investigating self-supervised pre-training for end-to-end speech
  translation,''
\newblock in {\em Proc. of the workshop on Self-supervision in Audio and Speech
  at the International Conference on Machine Learning (ICML)}, Virtual
  Conference, 2020.

\bibitem{Schneider2019wav2vecUP}
S.~Schneider, A.~Baevski, R.~Collobert, and M.~Auli,
\newblock ``wav2vec: Unsupervised pre-training for speech recognition,''
\newblock in {\em Proc. of INTERSPEECH}, Graz, Austria, 2019, pp. 3465--3469.

\bibitem{word2vec}
T.~Mikolov, I.~Sutskever, K.~Chen, G.~S. Corrado, and J.~Dean,
\newblock ``Distributed representations of words and phrases and their
  compositionality,''
\newblock in {\em Advances in Neural Information Processing Systems (NIPS)},
  Stateline, Nevada, USA, 2013, pp. 3111--3119.

\bibitem{Rodrigo2020}
R.~Pasti, F.~G. Vilasb{\^o}as, I.~R. Roque, and L.~N. de~Castro,
\newblock ``A sensitivity and performance analysis of {Word2Vec} applied to
  emotion state classification using a deep neural architecture,''
\newblock in {\em Proc. of the Distributed Computing and Artificial
  Intelligence Conference (DCAI)}, Ávila, Spain, 2019, pp. 199--206".

\bibitem{pustejovsky1996lexical}
A.P.C.S.J. Pustejovsky, B.~Pustejovsky, J.~Pustejovsky, and B.~Boguraev,
\newblock {\em Lexical Semantics: The Problem of Polysemy},
\newblock Clarendon paperbacks. Clarendon Press, 1996.

\bibitem{zhang2019ernie}
Z.~Zhang, X.~Han, Z.~Liu, X.~Jiang, M.~Sun, et~al.,
\newblock ``{ERNIE}: Enhanced language representation with informative
  entities,''
\newblock in {\em Proc. of ACL}, Florence, Italy, 2019, pp. 1441--1451.

\bibitem{martin-etal-2020-camembert}
L.~Martin, B.~Muller, P.~J. Ortiz~Su{\'a}rez, Y.~Dupont, L.~Romary, et~al.,
\newblock ``{CamemBERT}: a tasty {F}rench language model,''
\newblock in {\em Proc. of ACL}, Virtual Conference, 2020, pp. 7203--7219.

\bibitem{le2020flaubert}
H.~Le, L.~Vial, J.~Frej, V.~Segonne, M.~Coavoux, et~al.,
\newblock ``{FlauBERT}: Unsupervised language model pre-training for
  {F}rench,''
\newblock in {\em Proc. of Language Resources and Evaluation Conference
  (LREC)}, Virtual Conference, 2020, pp. 2479--2490.

\bibitem{Jing2018ProminenceFE}
S.~Jing, X.~Mao, and L.~Chen,
\newblock ``Prominence features: Effective emotional features for speech
  emotion recognition,''
\newblock {\em Digital Signal Processing}, vol. 72, pp. 216--231, 2018.

\bibitem{EGMAPS}
F.~Eyben, K.~Scherer, B.~Schuller, J.~Sundberg, E.~Andr{\'e}, et~al.,
\newblock ``The {Geneva} minimalistic acoustic parameter set {(GeMAPS)} for
  voice research and affective computing,''
\newblock {\em IEEE transactions on affective computing}, vol. 7, no. 2, pp.
  190--202, 2016.

\bibitem{schuller2013interspeech}
B.~Schuller, S.~Steidl, A.~Batliner, A.~Vinciarelli, K.~Scherer, et~al.,
\newblock ``The {INTERSPEECH} 2013 computational paralinguistics challenge:
  Social signals, conflict, emotion, autism,''
\newblock in {\em Proc. of INTERSPEECH}, Lyon, France, 2013, pp. 148--152.

\bibitem{chi2020audio}
P.~Chi, P.~Chung, T.~Wu, C.~Hsieh, S.~Li, et~al.,
\newblock ``Audio {AlBERT}: A lite {BERT} for self-supervised learning of audio
  representation,''
\newblock in {\em Pre-print on arXiv/2005.08575}, 2020.

\bibitem{AVEC}
F.~Ringeval, B.~Schuller, M.~Valstar, R.~Cowie, H.~Kaya, et~al.,
\newblock ``{AVEC} 2018 workshop and challenge: Bipolar disorder and
  cross-cultural affect recognition,''
\newblock in {\em Proc. of the Audio/Visual Emotion Challenge and Workshop},
  Beijing, China, 2018, pp. 3--13.

\bibitem{AVEC2019}
F.~Ringeval, B.~Schuller, M.~Valstar, N.~Cummins, R.~Cowie, et~al.,
\newblock ``{AVEC} 2019 workshop and challenge: State-of-mind, detecting
  depression with {AI}, and cross-cultural affect recognition,''
\newblock in {\em Proc. of the Audio/Visual Emotion Challenge and Workshop},
  Nice, France, 2019, pp. 3--12.

\bibitem{macary2020}
M.~Macary, M.~Lebourdais, M.~Tahon, Y.~Est{\`e}ve, and A.~Rousseau,
\newblock ``Multi-corpus experiment on continuous speech emotion recognition:
  convolution or recurrence?,''
\newblock in {\em Proc. of Conference on Speech and Computer (SPECOM)}, Virtual
  Conference, 2020.

\bibitem{OPEN_SMILE}
F.~Eyben, M.~W\"{o}llmer, and B.~Schuller,
\newblock ``{openSMILE} -- the munich versatile and fast open-source audio
  feature extractor,''
\newblock in {\em Proc. of the ACM Multimedia International Conference},
  Savannah,Georgia,USA, 2010, pp. 1459--1462.

\bibitem{gensim}
R.~{\v R}eh{\r u}{\v r}ek and P.~Sojka,
\newblock ``Software framework for topic modelling with large corpora,''
\newblock in {\em {Proc. of the LREC Workshop on New Challenges for NLP
  Frameworks}}, Valletta, Malta, 2010, pp. 45--50.

\bibitem{liu2019roberta}
Y.~Liu, M.~Ott, N.~Goyal, J.~Du, M.~Joshi, et~al.,
\newblock ``{RoBERTa}: A robustly optimized {BERT} pretraining approach,''
\newblock in {\em Pre-print on arXiv/1907.11692}, 2019.

\bibitem{ortizsuarez2019}
P.~J Ortiz~Su{\'a}rez, B.~Sagot, and L.~Romary,
\newblock ``Asynchronous pipeline for processing huge corpora on medium to low
  resource infrastructures,''
\newblock in {\em {Workshop on the Challenges in the Management of Large
  Corpora (CMLC-7)}}, Cardiff, United Kingdom, 2019, pp. 9--16.

\bibitem{basepaper}
M.~Schmitt, N.~Cummins, and B.~W. Schuller,
\newblock ``Continuous emotion recognition in speech - do we need
  recurrence?,''
\newblock in {\em Proc. of INTERSPEECH}, Graz, Austria, 2019, pp. 2808--2812.

\bibitem{CCC}
L.I-Kuei Lin,
\newblock ``A concordance correlation coefficient to evaluate
  reproducibility,''
\newblock {\em Biometrics}, vol. 45, no. 1, pp. 255--268, 1989.

\bibitem{pytorch}
A.~Paszke, S.~Gross, F.~Massa, A.~Lerer, J.~Bradbury, et~al.,
\newblock ``{PyTorch}: An imperative style, high-performance deep learning
  library,''
\newblock in {\em Advances in Neural Information Processing Systems (NIPS)},
  Vancouver, Canada, 2019, pp. 8024--8035.

\end{thebibliography}

\end{document}